\documentclass[conference]{IEEEtran}
\IEEEoverridecommandlockouts
\usepackage{cite}
\usepackage{amsmath,amssymb,amsfonts}
\usepackage{algorithmic}
\usepackage{graphicx}
\usepackage{textcomp}
\usepackage{multirow}
\usepackage{array}
\usepackage{xcolor}
\def\BibTeX{{\rm B\kern-.05em{\sc i\kern-.025em b}\kern-.08em
    T\kern-.1667em\lower.7ex\hbox{E}\kern-.125emX}}
\begin{document}

\title{A Cognitive Architecture for Shared Autonomy in AUV Operations\\
}

\author{Niamh Ellis$^{1,2}$, Thi Tran$^{2}$, Ignacio Carlucho$^{2}$ and Yvan R. Petillot$^{2}$
\thanks{$^{1}$School of Informatics, University of Edinburgh, Edinburgh, UK}%
\thanks{$^{2}$School of Engineering and Physical Sciences, Heriot-Watt University, Edinburgh, UK}%
\thanks{Niamh Ellis is the corresponding author. e-mail: nmfe2000@hw.ac.uk.}
}

\maketitle

\begin{abstract}
Operators remain essential to Remotely Operated Vehicle (ROV) operation, yet often suffer from low situational awareness and high workload, both of which negatively affect safety. 
This paper presents a cognitive architecture consisting of an ontology and multiple Large Language Models (LLMs) to assist the operator at all stages of the mission. Each LLM is grounded with domain-specific information from the ontology and given a simple role to create a system that can support the operator at all stages of an operation. We are aiming to prove that using the two together will allow decisions to be grounded in the relevant domain knowledge, but also benefit from the reasoning capabilities of the LLM.  Our framework determines if a mission is possible for a given Unmanned Underwater Vehicle (UUV), performs mission planning, and executes a given mission in simulation. The operator can be involved in planning and execution, ensuring the resulting plan is valid and that the vehicle behaves safely during execution. We compare different LLMs, Llama3, GPT-OSS, and Qwen2.5, to determine which are best suited to the different roles within our framework. We find that GPT-OSS performs best for feasibility assessment, planning, and execution, while Qwen2.5 is best suited to identifying mission types from natural language input. 
\end{abstract}


\section{Introduction}

Operators are essential to Remotely Operated Vehicle (ROV) missions.
However, operators often suffer from low situational awareness \cite{ho_human_2011} and a high workload, which affects safety and can increase the risk of mission-critical errors.  While Autonomous Underwater Vehicles (AUVs) could help alleviate some operator-related issues, full embedded autonomy remains a distant goal, and humans are still required in an oversight role. In these cases, shared-autonomy frameworks in which operator trust is considered from the outset are crucial to safe and effective operation. 

Cognitive architectures integrating domain knowledge and reasoning, have been developed in the past as a way to work alongside users in shared-autonomy frameworks. However, existing ontology-based approaches can struggle with representing implicit domain knowledge and performing logical reasoning at a human level \cite{zhou2024enhancing}. This can limit their use in more complex scenarios \cite{zhou2024enhancing, qi2024safetycontrolservicerobots}, which suggests that ontologies alone may struggle to properly handle the dynamic marine environment as well as the challenges that may arise during an operation.  

More recently, Large Language Models (LLMs) have appeared as a method capable of logical reasoning, making them able to understand complex problems and scenarios \cite{zhou2024enhancing}.
However, they can struggle with a deep understanding of the operating domain \cite{zhang2025surveygraphretrievalaugmentedgeneration} and are prone to hallucinations \cite{qi2024safetycontrolservicerobots, leanza2025conceptbotenhancingrobotsautonomy}.  
We theorise that using the two together will allow decisions to be grounded in the relevant domain knowledge, but also benefit from the reasoning capabilities of the LLM.

\begin{figure}
    \centering
    \includegraphics[width=\linewidth]{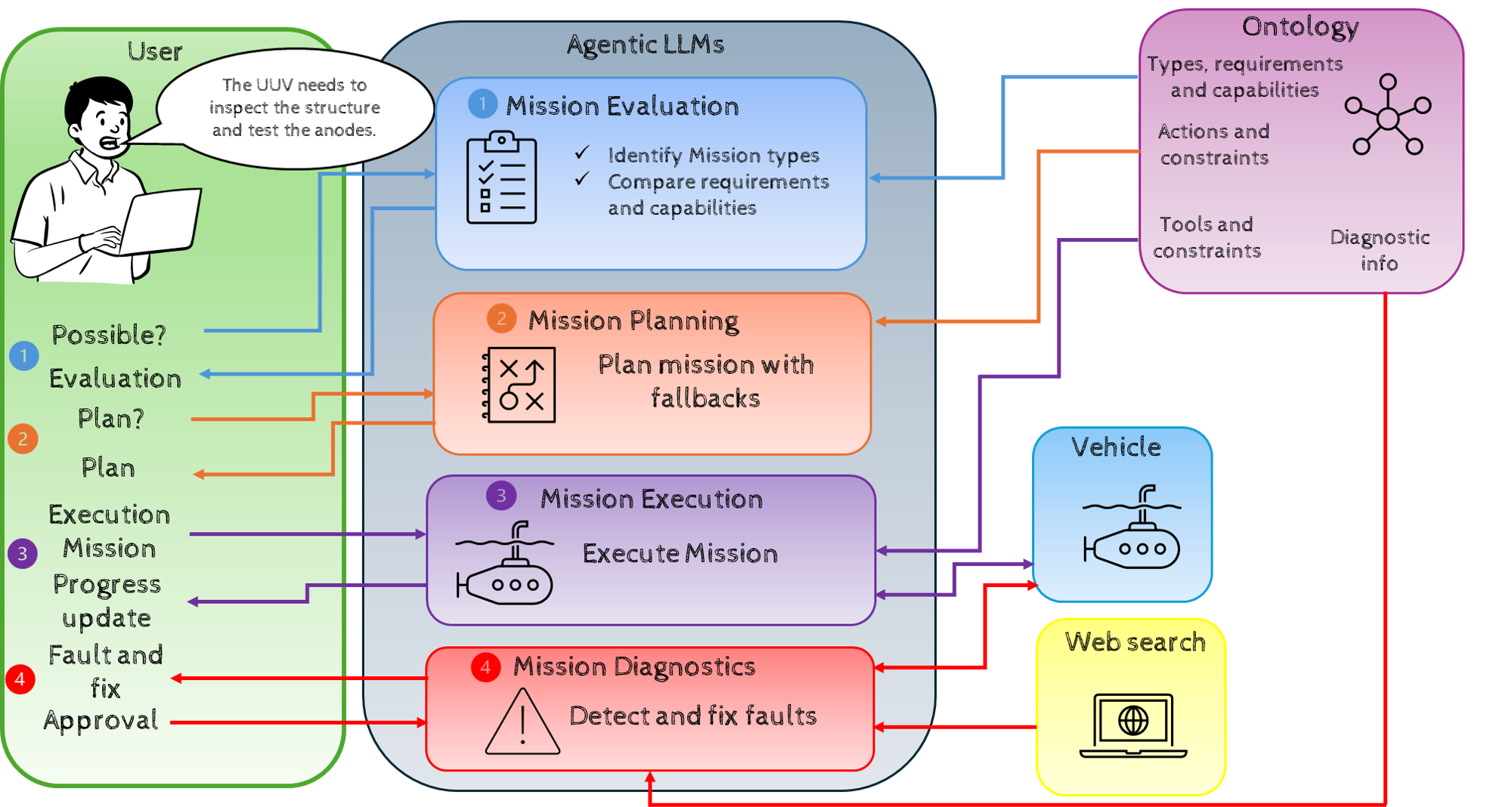}
    \caption{Diagram of the proposed Cognitive Framework showing communication between different components}
    \label{fig:CogFrame}
\end{figure}

In this work, we present a cognitive architecture (Fig. \ref{fig:CogFrame}) utilising agentic LLMs and an ontology as building blocks, incorporating domain-specific knowledge and the reasoning capabilities of an LLM. We theorise that using the LLMs and an ontology together will allow decisions to be grounded in the relevant domain knowledge, but also benefit from the reasoning capabilities of the LLM. Our proposed framework utilises multiple LLMs, each grounded in domain information from the ontology, to perform complex tasks. The complex task is broken down into a series of simple tasks for individual LLMs, each grounded with ontology information, the overall effect being reduced hallucinations. 
The complete system aims to aid the operator at multiple stages of the mission, from initial planning to execution and dealing with unexpected situations. 
We validate our framework using Stonefish \cite{stonefish2025} on an inspection mission with a simulated underwater vehicle. Results show that when using GPT-OSS \cite{openai2025gptoss120bgptoss20bmodel}, we achieve 100\% mission evaluation accuracy in different vehicle configurations. Additionally, we obtain 100\% valid plans with GPT-OSS for a simple inspection mission. Finally, we accomplish 100\% accurate mission execution in Stonefish with GPT-OSS.  

Our contributions are as follows:
\begin{itemize}
    \item A cognitive architecture, combining multiple role-specific LLMs with an ontology for grounding, to support shared autonomy at mission evaluation, planning and execution. 
    \item An evaluation of three LLMs (Llama3 \cite{grattafiori2024llama3herdmodels}, GPT-OSS and Qwen2.5\cite{qwen2025qwen25technicalreport}) for each role within the framework, identifying which models are best suited to mission evaluation, planning, and execution, respectively.
\end{itemize}

\section{Related Work}

\subsection{Knowledge representations}
Ontologies have been used for storing domain knowledge in marine autonomous systems. 
In \cite{SWARMS}, an ontology is used to allow different vehicles in a network to collect and share information. 
Unlike the work described in this paper, the system in \cite{SWARMS} does not utilise an LLM for its enhanced reasoning capabilities or for natural language communication with the user. 
In \cite{UnderwaterRule-BasedReasoner}, the authors expanded the ontology by adding SWRL rules, 
which gives the ontology increased reasoning capabilities (in the form of ``if-else-then" like statements) on top of the asserted knowledge.
However, in \cite{UnderwaterRule-BasedReasoner}, the focus is on their insertion and development of a rule-based reasoner and the usability of the SWRL rules for reasoning without any logical reasoning from an LLM. 
An ontology for use in fault recovery in the marine domain using deterministic and probabilistic information is developed in \cite{CoffeltBelief2022}. The stored deterministic information concerns relationships such as those between goals, tasks, capabilities, and components. While the probabilistic information focuses on the probability of faults, failures and potential fixes in the mission. 
A cognitive architecture is proposed in \cite{CoffletCognitive2023}, which is developed by combining multiple existing frameworks such as KnowRob and RoboSherlock. KnowRob is used as the knowledge service for the cognitive architecture. In order for KnowRob to be used in the marine domain, its ontologies must be extended. In our work, we interface our ontology with an LLM to allow additional natural language processing, in more than just goal-oriented tasks. We also allow the LLM to call actions in the environment by storing them in our ontology.    

While these knowledge representation-based approaches showcase the importance of domain knowledge, they lack logical reasoning. This form of reasoning, employed by humans, is likely essential in dynamic missions and is addressed by recent work with LLMs. 

\subsection{Large Language Models}


In recent years, LLMs have been used in many robotics domains. 
In \cite{buchholz2025collaborativereasoningframeworkanomaly}, a two LLM agent approach is used in diagnostics. The two LLMs are employed alongside a Vector database and a digital twin to aid operators in diagnosing causes of unexpected UUV behaviour. This system focuses on determining how to fix unexpected behaviour, unlike the framework proposed here, which first aims to determine if the UUV is capable of performing the mission and then to generate a plan and replan in the face of unexpected behaviour. This will give the user enhanced decision support in a shared-autonomy approach. 
An LLM is used in \cite{yang2024oceanplanhierarchicalplanningreplanning} for planning in the marine domain. An AUV is controlled based on natural language commands, while a vision-language model (VLM) converts images into text, which can be used to ground the LLM planner. Relevant domain knowledge is also provided in an XML file. Both this work and our work use relevant domain knowledge to ground the LLM. 
An LLM is used alongside ROS in the control of a robot via natural language commands in \cite{royce2025enablingnovelmissionoperations}. Natural language commands are used to allow users of different levels of experience to interact with and control robots. A Reasoning and Acting (ReAct) agent is implemented, which can understand these natural language commands and execute them on the robot. 
A part of our system performs similar behaviour with an LLM controlling the behaviour of a UUV in simulation. However, in our work, the LLM calls ROS 2 services and actions for implementing behaviour, which will allow us to monitor the progress of and cancel robot behaviours if necessary. 
%
%
However, 
LLMs alone are insufficient for many areas, including safety-critical marine operations. This has motivated research into combining them with structured knowledge representations. 

\subsection{Large Language Models and Knowledge Representations}

Work \cite{althobaiti2024llmsknowledgegraphscontribute} has used an LLM alongside a knowledge graph in a few-shot learning approach. The focus of this work is on safety in terms of interpreting and executing human natural language commands without violating safety considerations. Incorporating a safety mechanism is very important; however, the work in \cite{althobaiti2024llmsknowledgegraphscontribute} does not model capabilities or consider how they might be affected during an operation and how this might affect the outcome of said operation.  
An LLM and Knowledge graph system is developed in \cite{singh2025adaptbotcombiningllmknowledge} for a robotic agent working in a kitchen environment. In this work, the LLM is used to suggest a sequence of sub-tasks that can be used to perform a previously unseen task. The resulting set of sub-tasks may have some incorrect steps or make assumptions about objects and actions that the agent has. Therefore, a knowledge graph can be used to update the sequence of sub-tasks with domain-specific knowledge. This is similar to the planning component of our system; however, we preemptively ground the LLM with specific actions that our vehicle can perform. 
ConceptBot \cite{leanza2025conceptbotenhancingrobotsautonomy} is another framework utilising an LLM and a Knowledge graph in robotics. It is a modular robotics framework for generating achievable and safe plans in spite of natural language inputs containing ambiguity. The operating domain ConceptBot is used in is very different from the marine domain; additionally, there are gaps in the framework's knowledge graph in some domains, suggesting it would be unsuitable in our scenario. 
A knowledge graph and LLM-based system is developed in \cite{grimaldi2025advancingsharedmultiagentautonomy}. This work looks at both multi-agent and shared autonomy and utilises behaviour trees for mission execution. Unlike the work in this paper, our work utilises multiple LLMs in simple roles in an effort to reduce hallucinations.
In \cite{buchholz2025distributedaiagentscognitive}, a cognitive framework is presented using multiple LLMs and a vector database to store knowledge. Our work also utilises multiple LLMs; however, we focus on an end-to-end system for an operation instead of just the execution stage. 

\section{Methodology}
The designed system architecture (shown in Fig. \ref{fig:CogFrame}) is composed of: i) an ontology, ii) a set of agentic LLMs, iii) an LLM supervisor. The LLM agents work collaboratively to evaluate, plan and execute the mission, while the ontology provides grounding to the LLMs. Furthermore, we use ROS to channel the communication between the separate LLM agents, providing additional grounding. Below, we describe each element in more detail.

\subsection{Ontology} 
The ontology provides domain-relevant knowledge to the LLM agents, ensuring their responses are grounded in this information. 
Our ontology is specifically designed to store information relevant to the different components of our framework: i) mission evaluation, ii) mission planning, and iii) mission execution.

In terms of \textbf{mission evaluation}, we want our framework to determine if a given UUV is capable of performing a given user-input mission. Hence, within our ontology to aid the LLMs, we store information relating to vehicle components and possible missions.  An instance of an underwater vehicle can then be created to represent the specific vehicle being considered for the mission. This instance can then be related to all hardware, software and capabilities that it has using different relationships. 

Each mission type that a vehicle could be asked to perform, along with a brief description, is also represented in the ontology. These missions are also related to their required capabilities, for example, navigation. We make this decision for two reasons: first, to ground the system's response when determining whether the UUV can perform the given mission. Secondly, capabilities are considered vehicle independent (i.e. a vehicle can have different configurations of hardware, software and capabilities to have a given capability). We then utilise SWRL rules in the ontology to attribute a given capability to a vehicle for a specific component make-up. 

An example SWRL rule is given below: 
\begin{equation}
\begin{aligned}
\text{Platform}(?c) \land \text{hasCapability}(?c,
\text{Localisation}) \land {}        \\
\text{PathPlanningAlgorithm}(?d) 
\rightarrow \text{hasCapability}(?c,\text{PathPlanning})
\end{aligned}
\end{equation}
This denotes that an instance of the platform class has the capability of path planning if it has the capability of localisation (itself inferred from another SWRL rule) and a path planning algorithm on board. When the mission evaluation LLM queries the ontology for a user-defined mission, it retrieves both the required capabilities for that mission type and the inferred capabilities of the vehicle, allowing a grounded feasibility assessment without relying on the LLM to reason about hardware or software directly. 

Regarding \textbf{mission planning}, we want the LLM to select actions that a vehicle can perform and that are also relevant to the mission description. Therefore, we store actions that are a part of our mission types and their pre and post-conditions to aid the LLM in generating a valid plan for a given mission. This allows us to relate our actions to relevant software for implementing them with our vehicle. 
Additionally, we want the LLM to take into consideration any constraints posed by the vehicle or the environment on the plan. Consequently, we also store constraints relating to hardware, for instance, the maximum operating distance of a manipulator and also relevant environmental information, such as locations of permanent structures like wind turbines. We do this in an effort to ensure the plan is relevant to the vehicle and environment, for instance, directing the vehicle to a wind turbine or pipeline to begin an inspection or to a specific location at the end of a mission. 

Concerning \textbf{mission execution}, we want the LLM to have information about the specific software that can be used to perform mission actions. For this, we store ROS actions and their request parameters within the ontology; we denote some of these parameters as being provided by the LLM and some by the ontology. Parameters that are to be provided by the LLM are more mission-dependent; for instance, the LLM may need to adjust the standoff distance when inspecting a structure due to the environmental conditions. Parameters from the ontology, on the other hand, are static for the given mission, for example, the location of a structure to be inspected. 

\begin{figure}
    \centering
    \includegraphics[width=\linewidth]{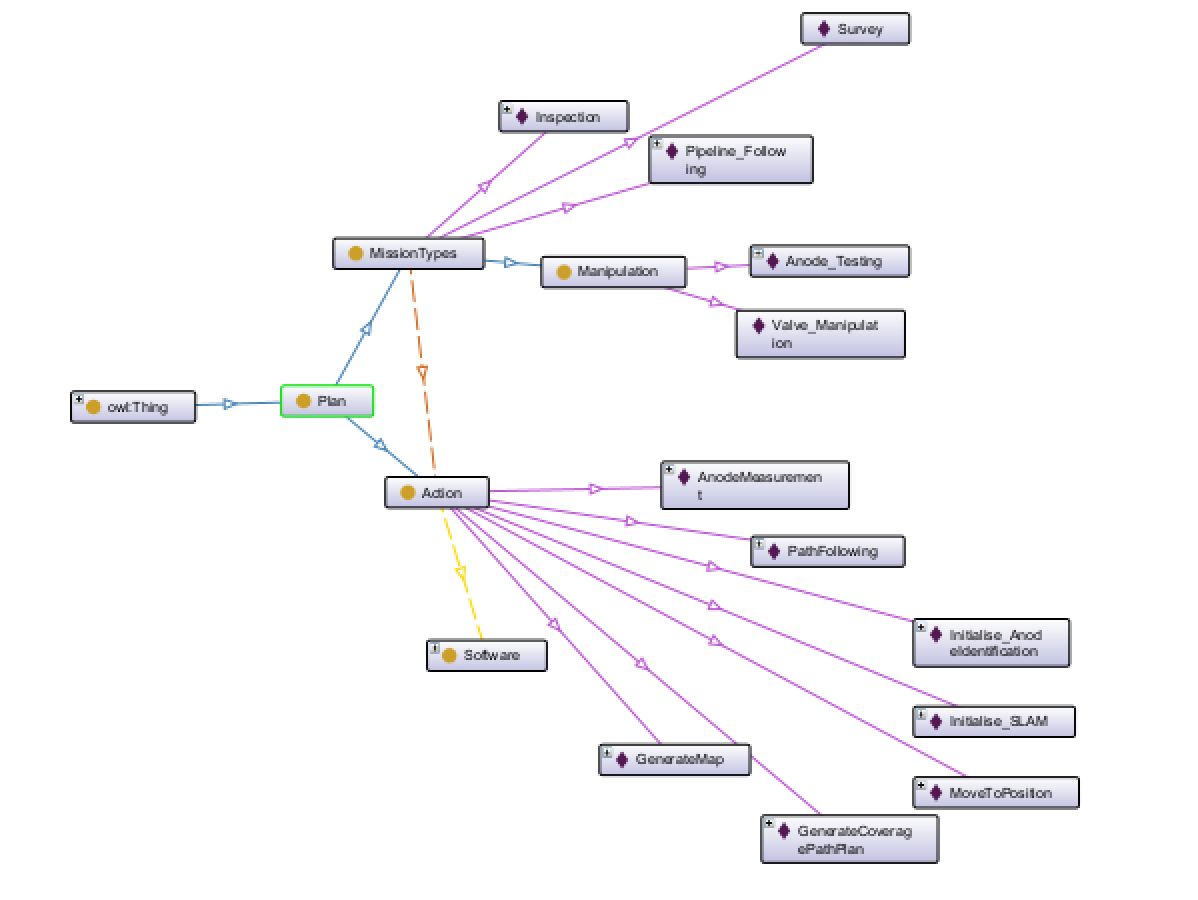}
    \caption{Snapshot of ontology showing connections between mission types and actions along with examples}
    \label{fig:ontologysnapswhot}
\end{figure}

Fig. \ref{fig:ontologysnapswhot} shows a snapshot of our ontology focused on the planning structure. Different colours of lines denote different types of relationships in the ontology; blue lines denote subclass relationships, while purple lines denote that an instance is a member of a specific class. The dashed lines denote different relationships possible between instances of the class. These are used to denote which actions are a part of a given mission and specific software that can be used to implement actions.

\subsection{LLM Agents}
Individual LLM agents are split into different roles, delivering simple tasks that, when combined, enable them to handle complex queries from the user. Each LLM within our framework is hosted locally via Ollama \cite{ollama}.

\textbf{Role 1: Mission Evaluation}, the first role of the agentic LLMs, is in determining if the UUV is capable of performing a given mission. Two LLMs are used in this, as shown in the mission evaluation box in Fig. \ref{fig:CogFrame}, the first of which is responsible for determining the types of missions in the user's mission description. We prompt the LLM with a system prompt containing the possible mission types from the ontology and the user's mission description and ask it to select the most relevant mission type(s). We use ROS to communicate these types to a second LLM node, which determines if the UUV can perform the mission. This LLM has a system prompt with information from the ontology concerning vehicle capabilities and required capabilities for the mission types. We instruct the LLM via this prompt to perform a multi-step process to determine mission possibility by comparing requirements with the vehicle's capabilities. The mission evaluation component can return 3 different results. These are fully possible, partially possible and impossible. For a mission to be fully possible, the given UUV instance, described in the ontology, must have all the required capabilities for a given mission. A partially possible mission corresponds to when the given UUV instance has the required capabilities for one or more parts of a multi-part mission, for example, for an inspection but not for anode testing. Finally, for a mission to be impossible, the given UUV instance must be missing the required capabilities for all parts of a mission

\textbf{Role 2: Mission Planning}, the second role the agentic LLMs perform is plan generation. For this, one LLM is used, which generates a plan collaboratively with the user. Information about actions, pre-conditions, and post-conditions of the identified mission types is fed into the LLM via its system prompt. After an initial plan has been generated, the user can ask the framework to improve on the plan before moving to the next stage. For example, a user may ask for additional steps to be added or for a specific part of a step to be revised or removed. Each subsequent plan can be additionally edited as required. The final plan can then be fed into the mission execution component so that the mission can be executed by the vehicle. 

\textbf{Roles 3 \& 4: Mission Execution and diagnostics}, The final two roles of the framework, execution and diagnostics, are performed largely in parallel. In the execution stage, an agentic LLM utilises a tool-based architecture to call actions within the environment. Here, the ontology is used to give information on how to invoke the required functionalities to deliver the mission. The operator is interfaced with this part of the framework in real time, allowing them to approve or edit tool calls as needed. Each tool call utilises a ROS action or service to fulfil the behaviour. This allows us to use tested implementations for behaviours instead of relying on the LLM to generate implementations of a given behaviour. We use ROS services to implement short-lived behaviours, such as planning a coverage path around a structure, and ROS actions for longer-lasting behaviours, such as path following, where we may require feedback or wish to cancel the action.
In the case of unexpected events, such as a piece of hardware failing, another LLM (LLM Diagnostics) is capable of detecting and analysing the fault. The LLM Diagnostics takes in information from the ontology and web searches to aid in its diagnosis. Through collaboration with the user, this component can suggest fixes or mitigations, including replanning the mission if necessary. Being able to cancel in-progress actions will be very beneficial for the mission diagnostics component.

\section{Results}\label{results}
We test our framework in Stonefish \cite{stonefish2025} for an inspection scenario using a model of an ROV  developed in our lab. The robot utilises ROS for communication, and is fitted with a stereo camera, a sonar, and an underwater manipulator, making it suited for an inspection in good environmental conditions.  
We evaluated the mission evaluation, mission planning, and mission execution components with the three different LLMs: i) Llama3, ii) Qwen2.5, and iii) GPT-OSS, and for different scenarios. All experiments were run on a ZBook Studio G10 with an i7-13700H processor and an RTX 4070 8GB graphics card. 

\subsection{Mission Evaluation Results}
We evaluate the \emph{Mission evaluation} module of our architecture.  
We instatiate three versions of our vehicle. The first has all the required capabilities for an inspection and an anode testing mission, including the manipulation and 3D stereo mapping. The second vehicle is missing essential capabilities for anode testing purposes, i.e., there is no manipulator onboard. However, it has all the requirements for an inspection mission. The third and final configuration is missing the required capabilities for both inspection and anode testing missions. 


\begin{table}[htbp]
\centering
\caption{Table of mission evaluation results for different configurations and different LLMs.}
\begin{tabular}{|l|l|c|c|c|}
\hline
\textbf{Configuration} & \textbf{Metric} & \textbf{Llama3} & \textbf{Qwen2.5} & \textbf{GPT-OSS} \\
\hline
Fully & Correct Feasibility & 60\% & 20\% & \textbf{100\%} \\
& w/o requirements & 100\% & 60\% & 100\% \\
\hline
Partially & Correct Feasibility & 80\% & 20\% & \textbf{100\%} \\
& w/o requirements & 40\% & 0\% & 40\% \\
\hline
Impossible & Correct Feasibility &100\% &100\%& \textbf{100\%} \\
& w/o requirements & 100\% & 100\% & 100\% \\
\hline
\end{tabular}
\label{table:MissionEvalResults}
\end{table}

The following mission description, ``\textit{An inspection of a structure to generate a 3D map and identify anodes which can then be tested}," is used for this test to ensure we can obtain all three possible results. We run each setup 5 times for each of the LLMs with the mission requirements from the ontology. To prove the value of the information in the ontology, we then run it a further 5 times, for each LLM, without the mission requirements from the ontology. The obtained results are shown in Table \ref{table:MissionEvalResults}. 
As can be seen from the Table, Qwen2.5 struggles to correctly identify fully or partially possible configurations, suggesting it is not suitable for this role. While Llama3 obtains better results, it still struggles to correctly identify the same configurations. 
Without the requirements from the ontology, we found that the LLMs often did well in the fully possible and impossible configurations but struggled in the partially possible configuration. In this configuration, each LLM can be overconfident and determine that the mission is fully possible, even though the vehicle is missing essential hardware, in this case, a manipulator. Furthermore, in cases when, without the required capabilities, the LLM correctly identifies that the given UUV can perform only part of the mission, it incorrectly identifies the missing component. This validates our use of an ontology to ground LLMs. 

We also evaluate the same LLMs for the role of identifying mission types in our framework; for this, the same mission description is fed in, and Table \ref{table:MissiontypeResults} is produced. 
This is an important part of our framework, as can be seen from the results in Table \ref{table:MissionEvalResults}, where the three LLMs can be overconfident in assigning Fully capable to a UUV that is only partially capable.

\begin{table}[t]
\begin{center}
\begin{tabular}{ |c|c| } 
 \hline
 \textbf{LLM}  & \textbf{Correct Types ID} \\ 
 \hline
Llama3 & 87\%\\  
 \hline
 Qwen2.5 & \textbf{100\%}  \\  
  \hline
  GPT-OSS & 87\% \\  
  \hline
\end{tabular}
\end{center}
\caption{Table of mission types results for different LLMs.}
\label{table:MissiontypeResults}
\end{table}

\subsection{Mission Planning Results}
In this section, we are evaluating the planning component of the framework. 
Here we are evaluating whether our framework is capable of developing valid plans. In this, we evaluate the following criteria:
\begin{itemize}
    \item Each step has its preconditions met by the step or steps before it, where necessary
    \item The final plan would successfully complete the mission objectives
    \item If the plan includes any redundant or contradictory steps 
\end{itemize}
We test this component using two different mission descriptions. The first of which is the same description used in the mission evaluation results. The second mission description is for a simpler mission, containing only one stage, and is ``\textit{An inspection of a structure to generate a 3D map.}"

We run 5 tests for each mission description. The results for the two mission descriptions are shown in Table \ref{table:MissionplanResults}.
\begin{table}[t]
\begin{center}
\begin{tabular}{ |>{\centering\arraybackslash}p{10mm}|>{\centering\arraybackslash}p{14mm}|>{\centering\arraybackslash}p{18mm}|>{\centering\arraybackslash}p{15mm}|>{\centering\arraybackslash}p{10mm}| } 
 \hline
 \textbf{LLM} & \textbf{Mission Type} & \textbf{Pre-conditions met} &\textbf{Successful completion} & \textbf{Optimal Plan}\\ 
 \hline
 \multirow{2}{4em}{Llama3} &Anode test  & 20\% & 0\% & 0\%\\  
 \cline{2-5}
 & Inspection & 100\% & 100\% &0\%\\ 
\hline
\multirow{3}{4em}{Qwen2.5} &Anode test  & 0\% & 0\% & 20\%\\  
 \cline{2-5}
 & Inspection & 100\% & 80\% &0\%\\ 
\hline
 \multirow{3}{4em}{GPT-OSS}  &Anode test  & 40\% & 40\% & 80\%\\  
 \cline{2-5}
 & Inspection & \textbf{100\%} & \textbf{100\%} &\textbf{60\%} \\ 
\hline
\end{tabular}
\end{center}
\caption{Table of mission planning results for different missions.}
\label{table:MissionplanResults}
\end{table} 
It can be seen from this table that overall GPT-OSS gets the strongest results in Anode testing missions, being the only LLM that produces plans that would successfully complete the anode inspection. Overall, however, the results for this mission type suggest that additional work is needed on this component of our framework.

In terms of the inspection mission, GPT-OSS performs the best overall, getting 100\% in both meeting pre-conditions and successful completion of the mission. While Llama3 gets comparative results in these metrics, it does generate multiple plans with redundant steps. 

\subsection{Mission Execution Results}
\begin{figure}[t]
    \centering
    \includegraphics[width=0.95\linewidth]{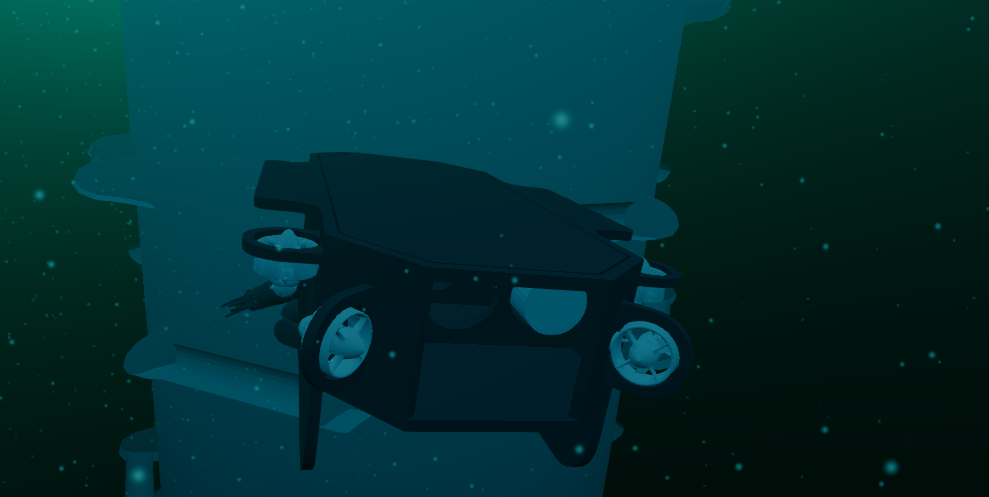}
    \caption{The ROV performing the inspection manuever in a wind turbine in Stonefish simulation environment.}
    \label{fig:stonefish}
\end{figure}

We also evaluate a ``perfect" mission with no faults during execution in Stonefish for an inspection mission. The vehicle and wind turbine being inspected are shown in Fig \ref{fig:stonefish}.
In this case, we use one plan generated by GPT-OSS, with no redundant or contradictory steps, for the mission description, ``\textit{An inspection of a structure to generate a 3D map}".
The following steps are in the plan:
\begin{enumerate}
    \item Initialise slam
    \item Generate Coverage Path plan
    \item Follow Path
    \item Move to a position near the surface 
\end{enumerate}

For this scenario, we develop ROS actions and services using existing methods to implement the required behaviour. These behaviours are:
\begin{itemize}
    \item Move to a 6 degree of freedom (DOF) position 
    \item SLAM for generating a map of the structure 
    \item Generation of a coverage path plan for the structure 
    \item Path following of a generated path
\end{itemize}
We evaluate the following criteria in terms of execution:
\begin{itemize}
    \item Number of actions called
    \item Number of incorrect actions called (i.e. action called at the wrong time)
    \item Number of times human correction is required (i.e. number of actions with the wrong name or wrong or missing parameters or called at the wrong time)
\end{itemize}
While we attempted testing on all three LLMs as in the previous tests, we have omitted the Llama3 results for our paper, as this LLM consistently struggled to execute the mission. While attempting these tests, Llama3 often generated incorrectly formatted output, making parsing for action calls impossible, or frequently generated incorrect action calls despite repeated human feedback. This highlights that Llama3 is unsuited to this role in our framework.
For Qwen2.5 and GPT-OSS models, Table \ref{table:MissionExecutionResults} has been produced. 
\begin{table}[t]
\begin{center}
\begin{tabular}{ |c|>{\centering\arraybackslash}p{15mm}|>{\centering\arraybackslash}p{20mm}|>{\centering\arraybackslash}p{25mm}| } 
 \hline
 \textbf{LLM} & \textbf{Actions called} & \textbf{Incorrect Actions } & \textbf{Times human feedback required}\\ 
 \hline
Qwen2.5& 7.4 & 2.4 & 3.4\\  
\hline
 GPT-OSS  & \textbf{4} & \textbf{0} & \textbf{0}\\  
\hline
\end{tabular}
\end{center}
\caption{Table of mission planning results for different missions. Numbers show average values.}
\label{table:MissionExecutionResults}
\end{table}

As has been seen from Tables \ref{table:MissionEvalResults}, \ref{table:MissionplanResults} and \ref{table:MissionExecutionResults}, GPT-OSS  performs best in the roles of mission evaluation, planning and execution. We theorise that this is due to the model being a reasoning or thinking model. On the other hand, Qwen2.5 performs best in the mission types identification role; we theorise that this is due to GPT-OSS over-reasoning about the wording of the user's mission description. And therefore, determining that the mission encompasses an inspection mission where anodes are identified but not specifically tested. 

\section{Conclusion and Future Work}
In conclusion, our framework is possible of evaluating a vehicle for a given mission, generating a simple plan for said mission and executing it under perfect conditions. We have identified which of the three tested LLMs is best suited to each role within our framework. For identifying mission types within a user input, we have identified Qwen2.5 as the best suited. When it comes to classifying a vehicle's ability to complete a given mission, we have identified GPT-OSS as best suited. Regarding planning, GPT-OSS performs best and also performs best in mission execution.

Our future work will focus on real-world testing of our framework and further development of components, particularly the diagnostics component. Developing our diagnostics part will allow our framework to work for more complex scenarios, adapting in the event of failure. This allows us to test our system more thoroughly and determine its true value in unforeseen circumstances when compared with more traditional methods. 
Real-world testing will allow us to further prove and validate our approach by exposing it to more challenging conditions. Further, it would allow us to model a range of potential faults that could be encountered during a mission.

\nocite{*}
\bibliographystyle{IEEEtran}
\bibliography{References}

@article{ho_human_2011,
	title = {Human {Factors} {Issues} with {Operating} {Unmanned} {Underwater} {Vehicles}},
	volume = {55},
	issn = {2169-5067},
	url_not = {https://journals.sagepub.com/doi/abs/10.1177/1071181311551088},
	doi = {10.1177/1071181311551088},
	language = {en},
	number = {1},
	urldate = {2023-11-25},
	journal = {Proceedings of the Human Factors and Ergonomics Society Annual Meeting},
	author = {Ho, Geoffrey and Pavlovic, Nada and Arrabito, Robert},
	month = sep,
	year = {2011},
	note = {Publisher: SAGE Publications Inc},
	pages = {429--433},
	file_not = {SAGE PDF Full Text:files/4450/Ho et al. - 2011 - Human Factors Issues with Operating Unmanned Under.pdf:application/pdf},
}

@article{zhou2024enhancing,
  title={Enhancing intention prediction and interpretability in service robots with {LLM} and {KG}},
  author={Zhou, Jincao and Su, Xuezhong and Fu, Weiping and Lv, Yang and Liu, Bo},
  journal={Scientific Reports},
  volume={14},
  number={26999},
  year={2024},
  doi={10.1038/s41598-024-77916-3},
}

@misc{qi2024safetycontrolservicerobots,
      title={Safety {Control} of {Service Robots} with {LLM}s and {Embodied Knowledge Graphs}}, 
      author={Yong Qi and Gabriel Kyebambo and Siyuan Xie and Wei Shen and Shenghui Wang and Bitao Xie and Bin He and Zhipeng Wang and Shuo Jiang},
      year={2024},
      howpublished_not={\url{https://doi.org/10.48550/arXiv.2405.17846}},
      eprint={2405.17846},
      archivePrefix={arXiv},
      primaryClass={cs.RO},
      url={https://arxiv.org/abs/2405.17846}, 
}

@misc{buchholz2025collaborativereasoningframeworkanomaly,
      title={A Collaborative Reasoning Framework for Anomaly Diagnostics in Underwater Robotics}, 
      author={Markus Buchholz and Niamh Ellis and Rahaf Abu Hara  and Ignacio Carlucho and Yvan R. Petillot},
      howpublished_not={\url{https://doi.org/10.48550/arXiv.2511.03075}},
      note={{IROS} 2026},
      year={2025},
      eprint={2511.03075},
      archivePrefix={arXiv},
      primaryClass={cs.RO},
      url={https://arxiv.org/abs/2511.03075}, 
}

@misc{qwen2025qwen25technicalreport,
      title={Qwen2.5 Technical Report}, 
      author={{Qwen}},
      author_not={Qwen and : and An Yang and Baosong Yang and Beichen Zhang and Binyuan Hui and Bo Zheng and Bowen Yu and Chengyuan Li and Dayiheng Liu and Fei Huang and Haoran Wei and Huan Lin and Jian Yang and Jianhong Tu and Jianwei Zhang and Jianxin Yang and Jiaxi Yang and Jingren Zhou and Junyang Lin and Kai Dang and Keming Lu and Keqin Bao and Kexin Yang and Le Yu and Mei Li and Mingfeng Xue and Pei Zhang and Qin Zhu and Rui Men and Runji Lin and Tianhao Li and Tianyi Tang and Tingyu Xia and Xingzhang Ren and Xuancheng Ren and Yang Fan and Yang Su and Yichang Zhang and Yu Wan and Yuqiong Liu and Zeyu Cui and Zhenru Zhang and Zihan Qiu},
      year={2025},
      eprint={2412.15115},
      archivePrefix={arXiv},
      primaryClass={cs.CL},
      url={https://arxiv.org/abs/2412.15115}, 
}

@misc{grattafiori2024llama3herdmodels,
      title={The Llama 3 Herd of Models}, 
      author={{Llama Team AI @ Meta}}, 
      author_not={Aaron Grattafiori and Abhimanyu Dubey and Abhinav Jauhri and Abhinav Pandey and Abhishek Kadian and Ahmad Al-Dahle and Aiesha Letman and Akhil Mathur and Alan Schelten and Alex Vaughan and Amy Yang and Angela Fan and Anirudh Goyal and Anthony Hartshorn and Aobo Yang and Archi Mitra and Archie Sravankumar and Artem Korenev and Arthur Hinsvark and Arun Rao and Aston Zhang and Aurelien Rodriguez and Austen Gregerson and Ava Spataru and Baptiste Roziere and Bethany Biron and Binh Tang and Bobbie Chern and Charlotte Caucheteux and Chaya Nayak and Chloe Bi and Chris Marra and Chris McConnell and Christian Keller and Christophe Touret and Chunyang Wu and Corinne Wong and Cristian Canton Ferrer and Cyrus Nikolaidis and Damien Allonsius and Daniel Song and Danielle Pintz and Danny Livshits and Danny Wyatt and David Esiobu and Dhruv Choudhary and Dhruv Mahajan and Diego Garcia-Olano and Diego Perino and Dieuwke Hupkes and Egor Lakomkin and Ehab AlBadawy and Elina Lobanova and Emily Dinan and Eric Michael Smith and Filip Radenovic and Francisco Guzmán and Frank Zhang and Gabriel Synnaeve and Gabrielle Lee and Georgia Lewis Anderson and Govind Thattai and Graeme Nail and Gregoire Mialon and Guan Pang and Guillem Cucurell and Hailey Nguyen and Hannah Korevaar and Hu Xu and Hugo Touvron and Iliyan Zarov and Imanol Arrieta Ibarra and Isabel Kloumann and Ishan Misra and Ivan Evtimov and Jack Zhang and Jade Copet and Jaewon Lee and Jan Geffert and Jana Vranes and Jason Park and Jay Mahadeokar and Jeet Shah and Jelmer van der Linde and Jennifer Billock and Jenny Hong and Jenya Lee and Jeremy Fu and Jianfeng Chi and Jianyu Huang and Jiawen Liu and Jie Wang and Jiecao Yu and Joanna Bitton and Joe Spisak and Jongsoo Park and Joseph Rocca and Joshua Johnstun and Joshua Saxe and Junteng Jia and Kalyan Vasuden Alwala and Karthik Prasad and Kartikeya Upasani and Kate Plawiak and Ke Li and Kenneth Heafield and Kevin Stone and Khalid El-Arini and Krithika Iyer and Kshitiz Malik and Kuenley Chiu and Kunal Bhalla and Kushal Lakhotia and Lauren Rantala-Yeary and Laurens van der Maaten and Lawrence Chen and Liang Tan and Liz Jenkins and Louis Martin and Lovish Madaan and Lubo Malo and Lukas Blecher and Lukas Landzaat and Luke de Oliveira and Madeline Muzzi and Mahesh Pasupuleti and Mannat Singh and Manohar Paluri and Marcin Kardas and Maria Tsimpoukelli and Mathew Oldham and Mathieu Rita and Maya Pavlova and Melanie Kambadur and Mike Lewis and Min Si and Mitesh Kumar Singh and Mona Hassan and Naman Goyal and Narjes Torabi and Nikolay Bashlykov and Nikolay Bogoychev and Niladri Chatterji and Ning Zhang and Olivier Duchenne and Onur Çelebi and Patrick Alrassy and Pengchuan Zhang and Pengwei Li and Petar Vasic and Peter Weng and Prajjwal Bhargava and Pratik Dubal and Praveen Krishnan and Punit Singh Koura and Puxin Xu and Qing He and Qingxiao Dong and Ragavan Srinivasan and Raj Ganapathy and Ramon Calderer and Ricardo Silveira Cabral and Robert Stojnic and Roberta Raileanu and Rohan Maheswari and Rohit Girdhar and Rohit Patel and Romain Sauvestre and Ronnie Polidoro and Roshan Sumbaly and Ross Taylor and Ruan Silva and Rui Hou and Rui Wang and Saghar Hosseini and Sahana Chennabasappa and Sanjay Singh and Sean Bell and Seohyun Sonia Kim and Sergey Edunov and Shaoliang Nie and Sharan Narang and Sharath Raparthy and Sheng Shen and Shengye Wan and Shruti Bhosale and Shun Zhang and Simon Vandenhende and Soumya Batra and Spencer Whitman and Sten Sootla and Stephane Collot and Suchin Gururangan and Sydney Borodinsky and Tamar Herman and Tara Fowler and Tarek Sheasha and Thomas Georgiou and Thomas Scialom and Tobias Speckbacher and Todor Mihaylov and Tong Xiao and Ujjwal Karn and Vedanuj Goswami and Vibhor Gupta and Vignesh Ramanathan and Viktor Kerkez and Vincent Gonguet and Virginie Do and Vish Vogeti and Vítor Albiero and Vladan Petrovic and Weiwei Chu and Wenhan Xiong and Wenyin Fu and Whitney Meers and Xavier Martinet and Xiaodong Wang and Xiaofang Wang and Xiaoqing Ellen Tan and Xide Xia and Xinfeng Xie and Xuchao Jia and Xuewei Wang and Yaelle Goldschlag and Yashesh Gaur and Yasmine Babaei and Yi Wen and Yiwen Song and Yuchen Zhang and Yue Li and Yuning Mao and Zacharie Delpierre Coudert and Zheng Yan and Zhengxing Chen and Zoe Papakipos and Aaditya Singh and Aayushi Srivastava and Abha Jain and Adam Kelsey and Adam Shajnfeld and Adithya Gangidi and Adolfo Victoria and Ahuva Goldstand and Ajay Menon and Ajay Sharma and Alex Boesenberg and Alexei Baevski and Allie Feinstein and Amanda Kallet and Amit Sangani and Amos Teo and Anam Yunus and Andrei Lupu and Andres Alvarado and Andrew Caples and Andrew Gu and Andrew Ho and Andrew Poulton and Andrew Ryan and Ankit Ramchandani and Annie Dong and Annie Franco and Anuj Goyal and Aparajita Saraf and Arkabandhu Chowdhury and Ashley Gabriel and Ashwin Bharambe and Assaf Eisenman and Azadeh Yazdan and Beau James and Ben Maurer and Benjamin Leonhardi and Bernie Huang and Beth Loyd and Beto De Paola and Bhargavi Paranjape and Bing Liu and Bo Wu and Boyu Ni and Braden Hancock and Bram Wasti and Brandon Spence and Brani Stojkovic and Brian Gamido and Britt Montalvo and Carl Parker and Carly Burton and Catalina Mejia and Ce Liu and Changhan Wang and Changkyu Kim and Chao Zhou and Chester Hu and Ching-Hsiang Chu and Chris Cai and Chris Tindal and Christoph Feichtenhofer and Cynthia Gao and Damon Civin and Dana Beaty and Daniel Kreymer and Daniel Li and David Adkins and David Xu and Davide Testuggine and Delia David and Devi Parikh and Diana Liskovich and Didem Foss and Dingkang Wang and Duc Le and Dustin Holland and Edward Dowling and Eissa Jamil and Elaine Montgomery and Eleonora Presani and Emily Hahn and Emily Wood and Eric-Tuan Le and Erik Brinkman and Esteban Arcaute and Evan Dunbar and Evan Smothers and Fei Sun and Felix Kreuk and Feng Tian and Filippos Kokkinos and Firat Ozgenel and Francesco Caggioni and Frank Kanayet and Frank Seide and Gabriela Medina Florez and Gabriella Schwarz and Gada Badeer and Georgia Swee and Gil Halpern and Grant Herman and Grigory Sizov and Guangyi and Zhang and Guna Lakshminarayanan and Hakan Inan and Hamid Shojanazeri and Han Zou and Hannah Wang and Hanwen Zha and Haroun Habeeb and Harrison Rudolph and Helen Suk and Henry Aspegren and Hunter Goldman and Hongyuan Zhan and Ibrahim Damlaj and Igor Molybog and Igor Tufanov and Ilias Leontiadis and Irina-Elena Veliche and Itai Gat and Jake Weissman and James Geboski and James Kohli and Janice Lam and Japhet Asher and Jean-Baptiste Gaya and Jeff Marcus and Jeff Tang and Jennifer Chan and Jenny Zhen and Jeremy Reizenstein and Jeremy Teboul and Jessica Zhong and Jian Jin and Jingyi Yang and Joe Cummings and Jon Carvill and Jon Shepard and Jonathan McPhie and Jonathan Torres and Josh Ginsburg and Junjie Wang and Kai Wu and Kam Hou U and Karan Saxena and Kartikay Khandelwal and Katayoun Zand and Kathy Matosich and Kaushik Veeraraghavan and Kelly Michelena and Keqian Li and Kiran Jagadeesh and Kun Huang and Kunal Chawla and Kyle Huang and Lailin Chen and Lakshya Garg and Lavender A and Leandro Silva and Lee Bell and Lei Zhang and Liangpeng Guo and Licheng Yu and Liron Moshkovich and Luca Wehrstedt and Madian Khabsa and Manav Avalani and Manish Bhatt and Martynas Mankus and Matan Hasson and Matthew Lennie and Matthias Reso and Maxim Groshev and Maxim Naumov and Maya Lathi and Meghan Keneally and Miao Liu and Michael L. Seltzer and Michal Valko and Michelle Restrepo and Mihir Patel and Mik Vyatskov and Mikayel Samvelyan and Mike Clark and Mike Macey and Mike Wang and Miquel Jubert Hermoso and Mo Metanat and Mohammad Rastegari and Munish Bansal and Nandhini Santhanam and Natascha Parks and Natasha White and Navyata Bawa and Nayan Singhal and Nick Egebo and Nicolas Usunier and Nikhil Mehta and Nikolay Pavlovich Laptev and Ning Dong and Norman Cheng and Oleg Chernoguz and Olivia Hart and Omkar Salpekar and Ozlem Kalinli and Parkin Kent and Parth Parekh and Paul Saab and Pavan Balaji and Pedro Rittner and Philip Bontrager and Pierre Roux and Piotr Dollar and Polina Zvyagina and Prashant Ratanchandani and Pritish Yuvraj and Qian Liang and Rachad Alao and Rachel Rodriguez and Rafi Ayub and Raghotham Murthy and Raghu Nayani and Rahul Mitra and Rangaprabhu Parthasarathy and Raymond Li and Rebekkah Hogan and Robin Battey and Rocky Wang and Russ Howes and Ruty Rinott and Sachin Mehta and Sachin Siby and Sai Jayesh Bondu and Samyak Datta and Sara Chugh and Sara Hunt and Sargun Dhillon and Sasha Sidorov and Satadru Pan and Saurabh Mahajan and Saurabh Verma and Seiji Yamamoto and Sharadh Ramaswamy and Shaun Lindsay and Shaun Lindsay and Sheng Feng and Shenghao Lin and Shengxin Cindy Zha and Shishir Patil and Shiva Shankar and Shuqiang Zhang and Shuqiang Zhang and Sinong Wang and Sneha Agarwal and Soji Sajuyigbe and Soumith Chintala and Stephanie Max and Stephen Chen and Steve Kehoe and Steve Satterfield and Sudarshan Govindaprasad and Sumit Gupta and Summer Deng and Sungmin Cho and Sunny Virk and Suraj Subramanian and Sy Choudhury and Sydney Goldman and Tal Remez and Tamar Glaser and Tamara Best and Thilo Koehler and Thomas Robinson and Tianhe Li and Tianjun Zhang and Tim Matthews and Timothy Chou and Tzook Shaked and Varun Vontimitta and Victoria Ajayi and Victoria Montanez and Vijai Mohan and Vinay Satish Kumar and Vishal Mangla and Vlad Ionescu and Vlad Poenaru and Vlad Tiberiu Mihailescu and Vladimir Ivanov and Wei Li and Wenchen Wang and Wenwen Jiang and Wes Bouaziz and Will Constable and Xiaocheng Tang and Xiaojian Wu and Xiaolan Wang and Xilun Wu and Xinbo Gao and Yaniv Kleinman and Yanjun Chen and Ye Hu and Ye Jia and Ye Qi and Yenda Li and Yilin Zhang and Ying Zhang and Yossi Adi and Youngjin Nam and Yu and Wang and Yu Zhao and Yuchen Hao and Yundi Qian and Yunlu Li and Yuzi He and Zach Rait and Zachary DeVito and Zef Rosnbrick and Zhaoduo Wen and Zhenyu Yang and Zhiwei Zhao and Zhiyu Ma},
      year={2024},
      eprint={2407.21783},
      archivePrefix={arXiv},
      primaryClass={cs.AI},
      url={https://arxiv.org/abs/2407.21783}, 
}

@misc{openai2025gptoss120bgptoss20bmodel,
      title={gpt-oss-120b \& gpt-oss-20b Model Card}, 
      author={OpenAI},
      author_Not={OpenAI and : and Sandhini Agarwal and Lama Ahmad and Jason Ai and Sam Altman and Andy Applebaum and Edwin Arbus and Rahul K. Arora and Yu Bai and Bowen Baker and Haiming Bao and Boaz Barak and Ally Bennett and Tyler Bertao and Nivedita Brett and Eugene Brevdo and Greg Brockman and Sebastien Bubeck and Che Chang and Kai Chen and Mark Chen and Enoch Cheung and Aidan Clark and Dan Cook and Marat Dukhan and Casey Dvorak and Kevin Fives and Vlad Fomenko and Timur Garipov and Kristian Georgiev and Mia Glaese and Tarun Gogineni and Adam Goucher and Lukas Gross and Katia Gil Guzman and John Hallman and Jackie Hehir and Johannes Heidecke and Alec Helyar and Haitang Hu and Romain Huet and Jacob Huh and Saachi Jain and Zach Johnson and Chris Koch and Irina Kofman and Dominik Kundel and Jason Kwon and Volodymyr Kyrylov and Elaine Ya Le and Guillaume Leclerc and James Park Lennon and Scott Lessans and Mario Lezcano-Casado and Yuanzhi Li and Zhuohan Li and Ji Lin and Jordan Liss and Lily and Liu and Jiancheng Liu and Kevin Lu and Chris Lu and Zoran Martinovic and Lindsay McCallum and Josh McGrath and Scott McKinney and Aidan McLaughlin and Song Mei and Steve Mostovoy and Tong Mu and Gideon Myles and Alexander Neitz and Alex Nichol and Jakub Pachocki and Alex Paino and Dana Palmie and Ashley Pantuliano and Giambattista Parascandolo and Jongsoo Park and Leher Pathak and Carolina Paz and Ludovic Peran and Dmitry Pimenov and Michelle Pokrass and Elizabeth Proehl and Huida Qiu and Gaby Raila and Filippo Raso and Hongyu Ren and Kimmy Richardson and David Robinson and Bob Rotsted and Hadi Salman and Suvansh Sanjeev and Max Schwarzer and D. Sculley and Harshit Sikchi and Kendal Simon and Karan Singhal and Yang Song and Dane Stuckey and Zhiqing Sun and Philippe Tillet and Sam Toizer and Foivos Tsimpourlas and Nikhil Vyas and Eric Wallace and Xin Wang and Miles Wang and Olivia Watkins and Kevin Weil and Amy Wendling and Kevin Whinnery and Cedric Whitney and Hannah Wong and Lin Yang and Yu Yang and Michihiro Yasunaga and Kristen Ying and Wojciech Zaremba and Wenting Zhan and Cyril Zhang and Brian Zhang and Eddie Zhang and Shengjia Zhao},
      year={2025},
      eprint={2508.10925},
      archivePrefix={arXiv},
      primaryClass={cs.CL},
      url={https://arxiv.org/abs/2508.10925}, 
}

@Article{SWARMS,

AUTHOR = {Li, Xin and Bilbao, Sonia and Martín-Wanton, Tamara and Bastos, Joaquim and Rodriguez, Jonathan},

TITLE = {{SWARM}s Ontology: A Common Information Model for the Cooperation of Underwater Robots},

JOURNAL = {Sensors},

VOLUME = {17},

YEAR = {2017},

NUMBER = {3},

ARTICLE-NUMBER = {569},

URL_not = {https://www.mdpi.com/1424-8220/17/3/569},

PubMedID = {28287468},

ISSN = {1424-8220},

DOI = {10.3390/s17030569}

}

@misc{ollama,
    title={Ollama},
    author={Ollama},
    url={https://ollama.com/},
    note = {Accessed: 03/07/2026}
}

@Article{UnderwaterRule-BasedReasoner,

AUTHOR = {Zhai, Zhaoyu and Martínez Ortega, José-Fernán and Lucas Martínez, Néstor and Castillejo, Pedro},

TITLE = {{A Rule-Based Reasoner for Underwater Robots Using OWL and SWRL}},

JOURNAL = {Sensors},

VOLUME = {18},

YEAR = {2018},

NUMBER = {10},

ARTICLE-NUMBER = {3481},

URL_not = {https://www.mdpi.com/1424-8220/18/10/3481},

PubMedID = {30332798},

ISSN = {1424-8220},

DOI = {10.3390/s18103481}

}

@inproceedings{CoffeltBelief2022,
author = {Coffelt, Jeremy and Mohammadi Kashani, Mahya and Wasowski, Andrzej and Kampmann, Peter},
year = {2022},
month = {08},
pages = {},
booktitle={The Eighth Joint Ontology Workshops (JOWO’22)},
title = {Belief-based fault recovery for marine robotics}
}

@INPROCEEDINGS{CoffletCognitive2023,
  author={Coffelt, Jeremy Paul and Beetz, Michael and Kampmann, Peter},
  booktitle={OCEANS 2023 - Limerick}, 
  title={Towards a Cognitive Architecture for Marine Robots}, 
  year={2023},
  volume={},
  number={},
  pages={1-8},
  doi={10.1109/OCEANSLimerick52467.2023.10244274}}

@misc{yang2024oceanplanhierarchicalplanningreplanning,
      title={OceanPlan: Hierarchical Planning and Replanning for Natural Language AUV Piloting in Large-scale Unexplored Ocean Environments}, 
      author={Ruochu Yang and Fumin Zhang and Mengxue Hou},
      year={2024},
      howpublished_not={\url{https://doi.org/10.48550/arXiv.2403.15369}},
      eprint={2403.15369},
      archivePrefix={arXiv},
      primaryClass={cs.RO},
      url={https://arxiv.org/abs/2403.15369}, 
}

@INPROCEEDINGS{royce2025enablingnovelmissionoperations,
  author={Royce, Rob and Kaufmann, Marcel and Becktor, Jonathan and Moon, Sangwoo and Carpenter, Kalind and Pak, Kai and Towler, Amanda and Thakker, Rohan and Khattak, Shehryar},
  booktitle={2025 IEEE Aerospace Conference}, 
  title={Enabling Novel Mission Operations and Interactions with {ROSA}: The Robot Operating System Agent}, 
  year={2025},
  volume={},
  number={},
  pages={1-16},
  doi={10.1109/AERO63441.2025.11068426}}

@misc{althobaiti2024llmsknowledgegraphscontribute,
      title={{How Can LLMs and Knowledge Graphs Contribute to Robot Safety? A Few-Shot Learning Approach}}, 
      author={Abdulrahman Althobaiti and Angel Ayala and JingYing Gao and Ali Almutairi and Mohammad Deghat and Imran Razzak and Francisco Cruz},
      howpublished_not={\url{https://doi.org/10.48550/arXiv.2412.11387}},
      year={2024},
      eprint={2412.11387},
      archivePrefix={arXiv},
      primaryClass={cs.RO},
      url={https://arxiv.org/abs/2412.11387}, 
}

@misc{grimaldi2025advancingsharedmultiagentautonomy,
      title={{Advancing Shared and Multi-Agent Autonomy in Underwater Missions: Integrating Knowledge Graphs and Retrieval-Augmented Generation}}, 
      author={Michele Grimaldi and Carlo Cernicchiaro and Sebastian Realpe Rua and Alaaeddine El-Masri-El-Chaarani and Markus Buchholz and Loizos Michael and Pere Ridao Rodriguez and Ignacio Carlucho and Yvan R. Petillot},
      year={2025},
      eprint={2507.20370},
      archivePrefix={arXiv},
      primaryClass={cs.RO},
      url={https://arxiv.org/abs/2507.20370}, 
}

@misc{singh2025adaptbotcombiningllmknowledge,
      title={{AdaptBot: Combining LLM with Knowledge Graphs and Human Input for Generic-to-Specific Task Decomposition and Knowledge Refinement}}, 
      author={Shivam Singh and Karthik Swaminathan and Nabanita Dash and Ramandeep Singh and Snehasis Banerjee and Mohan Sridharan and Madhava Krishna},
      year={2025},
      howpublished_not={\url{https://doi.org/10.48550/arXiv.2502.02067}},
      note={Accepted to IEEE International Conference on Robotics and Automation (ICRA) 2025},
      eprint={2502.02067},
      archivePrefix={arXiv},
      primaryClass={cs.RO},
      url_not={https://arxiv.org/abs/2502.02067}, 
}

@misc{buchholz2025distributedaiagentscognitive,
      title={Distributed AI Agents for Cognitive Underwater Robot Autonomy}, 
      author={Markus Buchholz and Ignacio Carlucho and Michele Grimaldi and Yvan R. Petillot},
      year={2025},
      eprint={2507.23735},
      archivePrefix={arXiv},
      primaryClass={cs.RO},
      url={https://arxiv.org/abs/2507.23735}, 
}

@misc{zhang2025surveygraphretrievalaugmentedgeneration,
      title={A Survey of Graph Retrieval-Augmented Generation for Customized Large Language Models}, 
      author={Qinggang Zhang and Shengyuan Chen and Yuanchen Bei and Zheng Yuan and Huachi Zhou and Zijin Hong and Hao Chen and Yilin Xiao and Chuang Zhou and Junnan Dong and Yi Chang and Xiao Huang},
      year={2025},
      eprint={2501.13958},
      archivePrefix={arXiv},
      primaryClass={cs.CL},
      url={https://arxiv.org/abs/2501.13958}, 
}

@misc{leanza2025conceptbotenhancingrobotsautonomy,
      title={ConceptBot: Enhancing Robot's Autonomy through Task Decomposition with Large Language Models and Knowledge Graph}, 
      author={Alessandro Leanza and Angelo Moroncelli and Giuseppe Vizzari and Francesco Braghin and Loris Roveda and Blerina Spahiu},
      year={2025},
      eprint={2509.00570},
      archivePrefix={arXiv},
      primaryClass={cs.RO},
      url={https://arxiv.org/abs/2509.00570}, 
}

@INPROCEEDINGS{stonefish2025,
  author={Grimaldi, Michele and Cieślak, Patryk and Ochoa, Eduardo and Bharti, Vibhav and Rajani, Hayat and Carlucho, Ignacio and Koskinopoulou, Maria and Petillot, Yvan R. and Gracias, Nuno},
  booktitle={2025 IEEE International Conference on Robotics and Automation (ICRA)}, 
  title={Stonefish: Supporting Machine Learning Research in Marine Robotics}, 
  year={2025},
  volume={},
  number={},
  pages={1-7},
  doi={10.1109/ICRA55743.2025.11127421}}
\end{document}